\pdfoutput=1

\documentclass[letterpaper, 10 pt, conference]{ieeeconf} 
\IEEEoverridecommandlockouts                     
\usepackage{graphicx}
\usepackage{subcaption}
\usepackage{algorithm} 
\usepackage{float}

\usepackage{algpseudocode}  
\usepackage{amsmath} 
\usepackage{threeparttable}
\usepackage{CJK}
\usepackage{indentfirst}
\usepackage{amsmath}
\usepackage{cases}

\usepackage{amssymb}
\usepackage{bbding}
\usepackage{utfsym}
\usepackage{amsmath,amssymb}
\usepackage{mathrsfs}
\usepackage{multirow}
\usepackage{soul, color, xcolor}
\usepackage[table]{xcolor}
\soulregister{\cite}7 % 注册\cite命令
\soulregister{\citep}7 % 注册\citep命令
\soulregister{\citet}7 % 注册\citet命令
\soulregister{\ref}7 % 注册\ref命令
\soulregister{\pageref}7 % 注册\pageref命令
\usepackage{cuted}
\newcommand{\Render}{\operatorname{Render}}
\newcommand{\FoundationPose}{\operatorname{FoundationPose}}

\usepackage{booktabs} 
\usepackage[hidelinks]{hyperref}

\title{\LARGE \bf
ActivePose:  Active 6D Object Pose Estimation and Tracking   \\ for Robotic Manipulation
}

\author{Sheng Liu\,$^{1,5}$,
Zhe Li\,$^{2}$, 
Weiheng Wang\,$^{1}$, 
Han Sun\,$^{*}$$^{2}$,  \\
Heng Zhang\,$^{3}$, 
Hongpeng Chen\,$^{4}$, 
Yusen Qin\,$^{5}$, 
Arash Ajoudani\,$^{3}$, 
Yizhao Wang\,$^{\dagger}$$^{2}$
	\thanks{* Han Sun is the project leader.} 
        \thanks{† Yizhao Wang is the corresponding author.}
	\thanks{Authors Affiliation: $^{1}$Karlsruhe Institute of Technology, Germany,
    $^{2}$Shanghai Jiao Tong University, China, 
    $^{3}$Istituto Italiano di Tecnologia, Italy, 
    $^{4}$The Hong Kong Polytechnic University, Hong Kong,
    $^{5}$D-Robotics, China.
    } 
   }

\begin{document}
\maketitle

\begin{abstract}
Accurate 6-DoF object pose estimation and tracking are critical for reliable robotic manipulation. 
However, zero-shot methods can fail under viewpoint-induced ambiguities, and fixed-camera setups struggle when objects move or become occluded. 
We propose \textit{ActivePose}, a closed-loop system that (i) actively disambiguates zero-shot pose estimates and (ii) actively tracks poses during downstream manipulation.

For active pose estimation, we combine a vision--language model (VLM) with CAD-based ``robot imagination.'' Offline, we render CAD views, compute FoundationPose hypothesis entropy from rendered views, and build a geometry-aware prompt with low-entropy (unambiguous) and high-entropy (ambiguous) exemplars. Online, the system queries the VLM for the ambiguity probability of the current view; when ambiguous, it renders virtual views from an IK-feasible candidate set, scores each view by fusing VLM-predicted ambiguity and entropy-based uncertainty, and executes the selected Next-Best-View (NBV) for disambiguation.

For active pose tracking, we train a diffusion policy via imitation learning to generate receding-horizon camera trajectories that preserve target visibility and reduce pose-loss under object motion and occlusions. Experiments in simulation and on real robots demonstrate consistent improvements over classical baselines. %We release our code at \url{https://anonymous.4open.science/r/ActivePose/}.
\end{abstract}

\section{INTRODUCTION}

6D object pose estimation, which recovers an object's translation and rotation relative to the camera, plays a pivotal role in robotic manipulation tasks such as grasping \cite{choi2018learning, fang2020graspnet, zhang20216dof} and assembly \cite{malik2019advances}.
Previous works \cite{peng2019pvnet, zakharov2019dpod, li2019cdpn, cheng2021real, wang2021gdr, zhou20226} achieved strong performance by training on specific datasets.
More recently, zero-shot methods \cite{nguyen2024gigapose, wen2024foundationpose, lin2024sam} demonstrated the ability to estimate poses for novel objects using only their CAD models, without requiring additional real-world annotations.

Despite their strong performance, most existing methods assume that the observation from a single viewpoint contains enough cues to estimate a unique pose.
In practice, however, self-occlusion and inter-object occlusion frequently obscure distinctive regions, resulting in fundamentally ill-posed estimation problems.
This challenge is especially prevalent in industrial metal parts, where symmetric structures and textureless surfaces inherently reduce distinctive regions, leading to severely exacerbated pose ambiguity.

Since pose ambiguity is inevitable in practical scenarios, resolving it becomes critically important.
While physically reorienting the target object or removing occluders can resolve pose ambiguity, such direct interventions are often prohibited in tasks involving high-precision components or delicate instruments.
Instead, enabling cameras to actively adjust viewpoints, which mimics human observation, offers a more versatile solution.
\begin{figure}[htbp]
  \centering
  \includegraphics[scale=0.47]{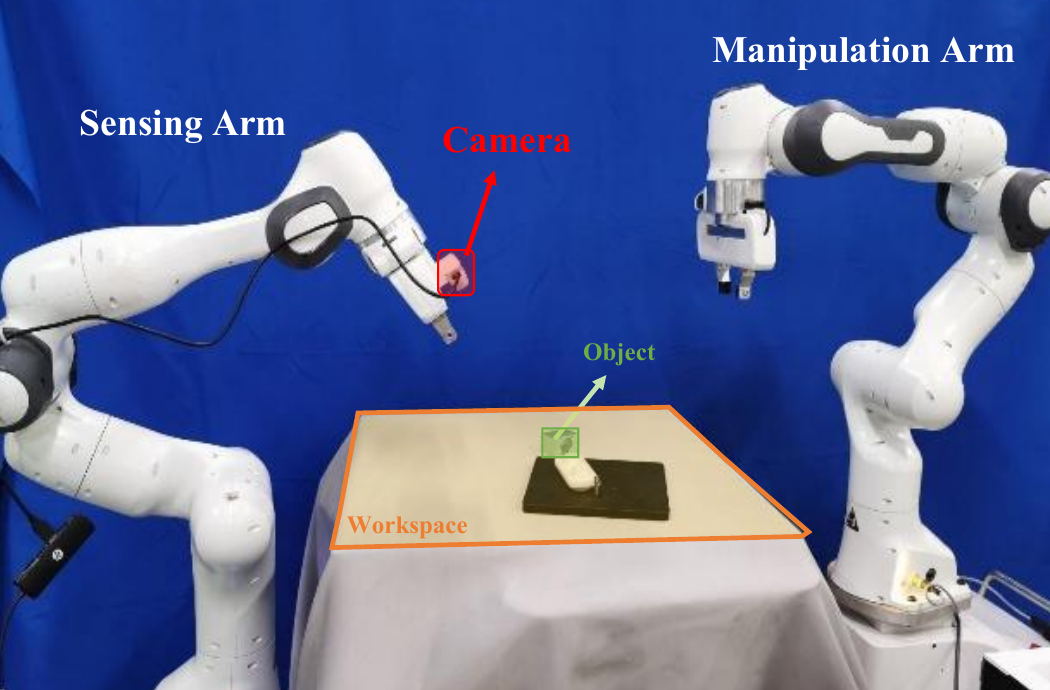}
  \caption{Dual-arm experimental setup. The left arm serves as the \emph{sensing arm} and carries a wrist-mounted RGB-D camera, while the right arm serves as the \emph{manipulation arm} equipped with a parallel-jaw gripper.}
  \label{workspace}
   %\vspace{-10pt}
\end{figure}

Although prior studies \cite{hofer2023hyperposepdf, murphy2021implicit} attempted to predict pose ambiguity, they not only depend on costly prior information, but they also fail to provide actionable solutions for ambiguity elimination. Several methods have trained neural networks to predict the Next-Best-View (NBV) for active pose estimation \cite{9340842,yang2025active6dposeestimation,Mizuno_2024}. Although these approaches show promise, they require extensive offline training.
Thus, determining optimal camera movements to mitigate pose ambiguity during operation remains an open and challenging problem.

With this motivation, we propose \textit{ActivePose}, a closed-loop framework for zero-shot 6D pose disambiguation and downstream pose tracking.
We ground a VLM with entropy-ranked CAD renderings to detect viewpoint-induced ambiguity and select NBVs from a small IK-feasible candidate set via renderer-based view synthesis. We further learn a diffusion-policy tracker to actively maintain visibility under motion and occlusions.

To our knowledge, ActivePose is among the first closed-loop frameworks that combines zero-shot ambiguity detection with feasible NBV selection for CAD-based novel-object pose estimation, and further integrates active tracking for downstream manipulation under motion and occlusions. 
For the benefit of the research community, we will release our code as open source.

In summary, our contributions are:
\begin{itemize}
    %\item A zero-shot ambiguity detector that grounds a VLM with entropy-ranked CAD renderings and outputs an ambiguity probability for the current view.
    %\item A NBV selection rule that ranks candidate views by combining VLM ambiguity with FoundationPose hypothesis entropy, enabling closed-loop pose disambiguation without training an NBV policy.
    \item A zero-shot active pose estimation module that detects viewpoint ambiguity and executes feasible NBV moves to disambiguate CAD-based 6D pose estimates in closed loop.
    \item A demonstration-trained diffusion-policy tracker that generates  camera trajectories to prevent pose-loss under motion and occlusions.
    \item Evaluation in simulation and on real dual-arm hardware, including an industrial peg-in-hole assembly case study and a runtime analysis of VLM latency.
\end{itemize}

\begin{figure*}
	\centering
	\includegraphics[width=1\linewidth]{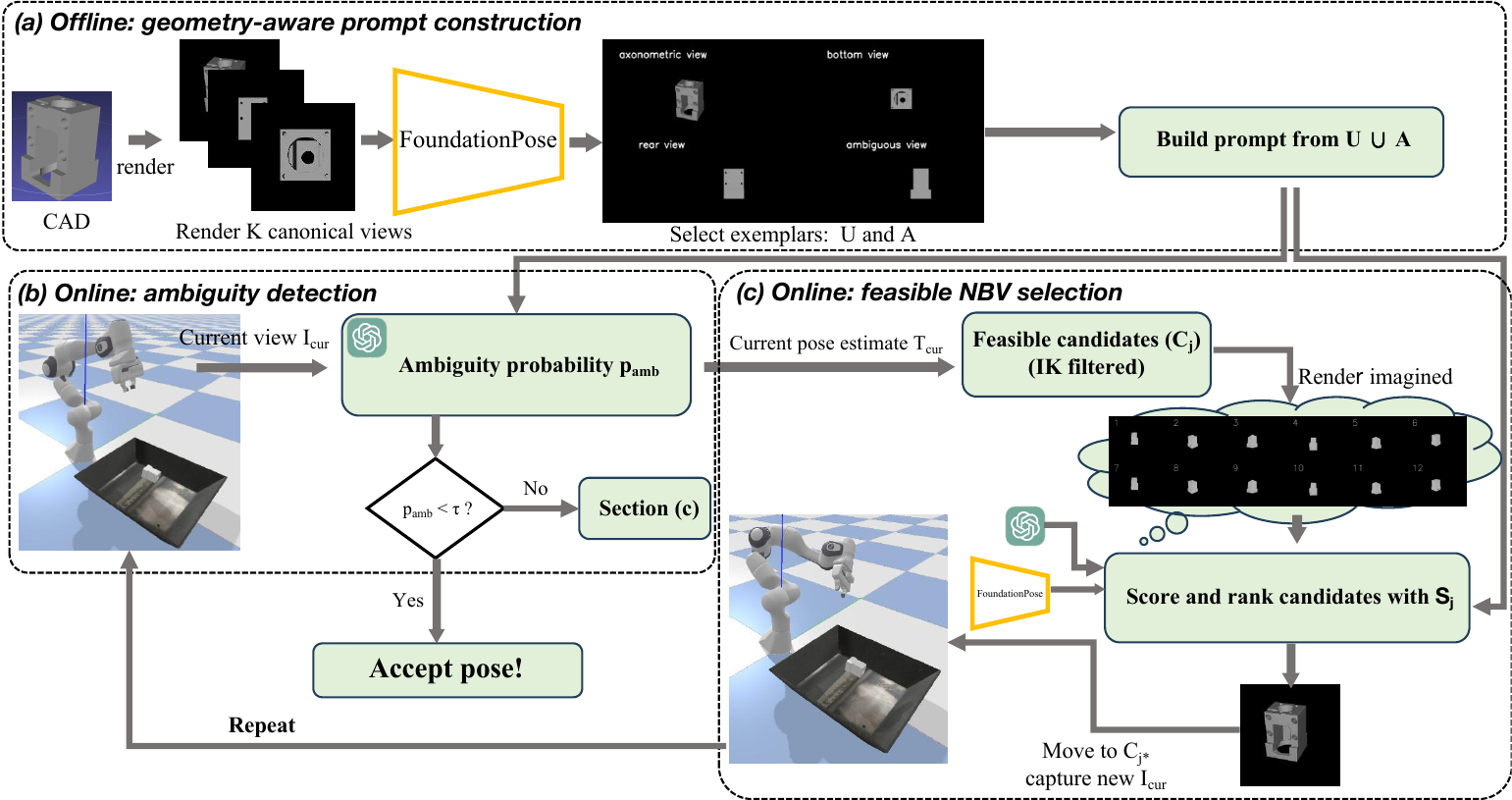}
	\caption{\textbf{Active pose estimation.} (a) Offline: render canonical CAD views, compute the hypothesis entropy of FoundationPose, and build a geometry-aware prompt from low-/high-entropy exemplars. (b) Online: compute VLM ambiguity $p_{\mathrm{amb}}$ for the current view and trigger disambiguation when $p_{\mathrm{amb}}>\tau$. (c) Rank feasible candidate views using rendered imagined observations and the fused score, execute the selected NBV, and repeat up to budget $L$.}

	\label{pipeline}
\end{figure*}

\section{RELATED WORK}

\subsection{6D Pose Estimation}
In recent years, 6D pose estimation methods based on a single RGB-D image using neural networks \cite{peng2019pvnet,zakharov2019dpod, li2019cdpn, park2019pix2pose,he2020pvn3ddeeppointwise3d,  9565319} have outperformed the classic approaches \cite{hodavn2020bop}. However, these methods did not consider the ambiguous pose from a specific viewpoint. To address this, active pose estimation techniques adjust the camera viewpoint to improve overall confidence—but they typically rely on object-specific training, hand-crafted heuristics, or expensive annotations \cite{9340842,yang2025active6dposeestimation,Mizuno_2024}. Differently, we target \emph{zero-shot} ambiguity detection and disambiguation for novel CAD objects: we ground a VLM with entropy-ranked CAD renders to assess geometric ambiguity, and select NBVs from a feasible candidate set via renderer-based view synthesis (``robot imagination'') in a closed loop.

\subsection{Vision-Language Models for Robotics}
The rapid development of VLMs has further advanced the field of robotics, enabling intelligent agents to interpret natural language commands and reason about visual scenes in a generalizable, task-independent manner \cite{shao2025largevlmbasedvisionlanguageactionmodels}. A key advantage of these basic models is that they have been pre-trained on large-scale datasets and can usually be directly deployed in robot systems as high-level planners \cite{driess2023palmeembodiedmultimodallanguage,liang2023codepolicieslanguagemodel,vemprala2023chatgptroboticsdesignprinciples} or low-level controllers \cite{li2024visionlanguagefoundationmodelseffective,wang2025robobertendtoendmultimodalrobotic,kim2024openvlaopensourcevisionlanguageactionmodel} without having to learn from scratch, which significantly reduces training costs and improves generalization capabilities. In our work, we use a VLM not as a task planner, but as a \emph{geometric ambiguity assessor}: grounded by geometry-aware exemplars, it outputs an ambiguity probability for the current view and helps rank candidate viewpoints for zero-shot active disambiguation.

\subsection{Diffusion Models for Robotics}
Diffusion models have recently emerged as a powerful paradigm for robot learning, particularly for manipulation, by modeling complex and multimodal action or trajectory distributions and generating smooth behaviors through iterative denoising \cite{wolf2025diffusionmodelsroboticmanipulation}. 
A representative approach, Diffusion Policy \cite{chi2024diffusionpolicyvisuomotorpolicy}, formulates visuomotor control as a conditional diffusion process that predicts action sequences and executes them in a receding-horizon loop, achieving strong performance across diverse manipulation benchmarks.

In contrast to prior diffusion-based manipulation policies that primarily generate end-effector motions for task execution, we use a diffusion policy as an \emph{active sensing} component: the policy generates receding-horizon camera trajectories that maintain target visibility and reduce pose-loss under object motion and occlusions during downstream manipulation.

%%%%%%%%%%%%%%%%%%%%%%%%%%%%%%%%%%%%%%%%%%%%%%%%%%

%\noindent\textbf{Summary.} Prior work either focuses on single-view zero-shot pose estimation without active disambiguation, or learns NBV policies with substantial offline training. ActivePose bridges this gap by providing a zero-shot, feedback-driven framework that detects viewpoint-induced ambiguity and selects feasible NBV for disambiguation, while further maintaining pose accuracy via active tracking during downstream manipulation.

\section{Problem Formulation}
We formulate active 6D object pose estimation and tracking as a two-stage closed-loop problem.
Given a CAD model, an initial camera pose, and a live RGB-D stream, the system first performs \emph{active pose estimation}: it detects viewpoint-induced ambiguity using a VLM grounded by geometry-aware exemplars together with FoundationPose hypothesis entropy, and, when ambiguous, moves the camera to a next-best view selected by ranking rendered candidates to obtain an unambiguous pose.
After obtaining this disambiguated pose, the system performs \emph{active pose tracking} during manipulation: a diffusion-policy tracker generates camera trajectories to preserve visibility and prevent pose-loss under object motion and occlusions.

\section{Method}
Our approach comprises two tightly integrated modules: an active pose estimation pipeline that resolves viewpoint-induced 6D pose ambiguities (Algorithm~\ref{alg:active-pose-estimation}), and an active pose tracking module that actively moves the camera to preserve visibility and prevent pose-loss under motion and occlusions (Algorithm~\ref{alg:active_pose_tracking}).

\subsection{Active Pose Estimation}
Our active pose estimation pipeline (see Fig.~\ref{pipeline}) comprises two stages: (i) \emph{offline} geometry-aware prompt construction and (ii) \emph{online} ambiguity detection with feasible NBV selection.

\textbf{Offline:} We pre-render $K$ canonical views $\{R_k\}_{k=1}^{K}$ by sampling camera poses on a tabletop-facing hemisphere around the object (fixed standoff distance $d$, camera optical axis pointing to the object center), and reuse these renders to construct a geometry-aware prompt that remains fixed for all online queries of the same object. These canonical views are used only for prompt construction and do not need to be kinematically feasible for the robot.
For each rendered view $R_k$, we invoke FoundationPose \cite{ wen2024foundationpose} to obtain the top-$N$ \emph{6D} pose hypotheses
$\{\widehat T_i\}_{i=1}^{N}$ with associated confidence weights $\{w_i\}_{i=1}^{N}$.
We normalize the weights to form a discrete distribution $p_i = \frac{w_i}{\sum_{j=1}^{N} w_j}$ and compute the
\emph{normalized} Shannon entropy
\begin{equation}
\bar H(R_k)
=
\frac{-\sum_{i=1}^{N} p_i \log p_i}{\log N}\in[0,1].
\label{eq:pose_entropy}
\end{equation}

This serves as a geometry-driven uncertainty score that is comparable across views.
We then select two small subsets: $U$ as the $K_u$ lowest-entropy \emph{unambiguous} views and $A$ as the $K_a$ highest-entropy \emph{ambiguous} views.
The geometry-aware prompt is constructed by aggregating the images in $U \cup A$ with a fixed instruction template.

\begin{algorithm}[t]
\caption{Active Pose Estimation}
\label{alg:active-pose-estimation}
\begin{algorithmic}[1]
\Require CAD model; ambiguity threshold $\tau$; fusion weight $\lambda$; NBV budget $L$;
candidate-view generator $\mathcal{G}(\cdot)$ with IK filtering
\Ensure final 6D object pose in camera frame ${}^{c}\!T_{o}$

\State \textbf{Offline:}
\State Render $K$ canonical views $\{R_k\}_{k=1}^{K}$
\State Compute normalized pose entropies $\{\bar H_k\}_{k=1}^{K}$ from FoundationPose hypotheses (Eq.~\eqref{eq:pose_entropy})
\State $U \gets$ $K_u$ lowest-entropy views; \quad $A \gets$ $K_a$ highest-entropy views
\State $\text{prompt} \gets \textsc{BuildPrompt}(U,A)$

\State \textbf{Online:}
\State Acquire $I_{\mathrm{cur}}$ at current camera pose ${}^{b}\!T_c$
\State $T_{\mathrm{cur}} \gets \FoundationPose(I_{\mathrm{cur}})$
\For{$\ell=0$ \textbf{to} $L$}
  \State $p_{\mathrm{amb}} \gets \textsc{AmbiguityProb}(I_{\mathrm{cur}},\text{prompt})$ (Eq.~\eqref{eq:p})
  \If{$p_{\mathrm{amb}} < \tau$ \textbf{or} $\ell = L$}
    \State \textbf{break}
  \EndIf
  %\State $\{C_j\}_{j=1}^{M} \gets \mathcal{G}({}^{b}\!T_c, T_{\mathrm{cur}})$
  \State ${}^{b}\!T_o \gets {}^{b}\!T_c \cdot T_{\mathrm{cur}}$
  \State $\{C_j\}_{j=1}^{M} \gets \mathcal{G}({}^{b}\!T_c, {}^{b}\!T_o)$
  \For{$j=1,\dots,M$}
    \State $\hat I_j \gets \Render(C_j)$
    \State $p_{\mathrm{amb},j} \gets \textsc{AmbiguityProb}(\hat I_j,\text{prompt})$ (Eq.~\eqref{eq:p})
    \State $\bar H_j \gets \textsc{PoseEntropy}(\hat I_j)$ (Eq.~\eqref{eq:pose_entropy})
\State $S_j \gets \lambda \bar H_j + (1-\lambda)p_{\mathrm{amb},j}$
  \EndFor
  \State $j^* \gets \arg\min_j S_j$
  \State Move camera to $C_{j^*}$ and acquire new real image $I_{\mathrm{cur}}$
  \State ${}^{b}\!T_c \gets C_{j^*}$
  \State $T_{\mathrm{cur}} \gets \FoundationPose(I_{\mathrm{cur}})$
\EndFor
\State ${}^{c}\!T_{o} \gets T_{\mathrm{cur}}$
\end{algorithmic}
\end{algorithm}

\textbf{Online:}
Given the current observation $I_{\mathrm{cur}}$, we estimate an initial pose
$T_{\mathrm{cur}}=\FoundationPose(I_{\mathrm{cur}})$ and compute an ambiguity probability
$p_{\mathrm{amb}}$ using a fixed, per-object prompt (constructed offline) and constrained
\{\texttt{Yes}, \texttt{No}\} log-probabilities from the VLM:
\begin{equation}
\begin{aligned}
p_{\mathrm{amb}}
&= P_{\mathrm{VLM}}(\texttt{Yes}\mid I_{\mathrm{cur}}, \text{prompt}) \\
&= \frac{\exp(\ell_{\mathrm{Yes}})}
{\exp(\ell_{\mathrm{Yes}})+\exp(\ell_{\mathrm{No}})} .
\end{aligned}
\label{eq:p}
\end{equation}
If $p_{\mathrm{amb}}<\tau$, we directly output $T_{\mathrm{cur}}$; otherwise we iteratively select a feasible NBV within a small budget $L$. A larger $\tau$ makes disambiguation less frequent (only highly ambiguous views trigger NBV), while a smaller $\tau$ triggers NBV more aggressively.
Given the current camera pose ${}^{b}\!T_c$ and the current pose estimate $T_{\mathrm{cur}}$, the generator $\mathcal{G}$ samples $M$ \emph{local} candidate camera poses around the object and retains only IK-feasible candidates.
For each feasible candidate $C_j$, we render an imagined view $\hat I_j$ and compute $p_{\mathrm{amb},j}=P_{\mathrm{VLM}}(\texttt{Yes}\mid \hat I_j,\text{prompt})$ with the same prompt.
We score candidates by $S_j=\lambda\,\bar H(\hat I_j) + (1-\lambda)\,p_{\mathrm{amb},j}$, then move the camera to the best-scoring candidate view, and acquire a new real observation.
We repeat this process until the newly acquired image yields $p_{\mathrm{amb}}<\tau$ or the budget $L$ is exhausted (Algorithm~\ref{alg:active-pose-estimation}); at termination, we return the current estimate as the final pose ${}^{c}\!T_{o}$.

\subsection{Active Pose Tracking}
\begin{figure}[t]
  \centering
  \includegraphics[scale=0.63]{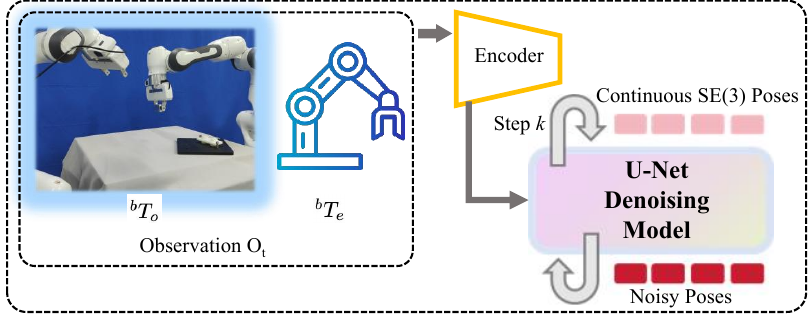}
  \caption{\textbf{Active pose tracking.} The policy encodes the current observation $O_t$, denoises over $K_d$ reverse-diffusion steps to generate a horizon of continuous SE(3) poses, and executes the last $k_h$ poses in a receding-horizon loop.}
  \label{fig:active_pose_tracking}
\end{figure}

\begin{algorithm}[ht]
\caption{Active Pose Tracking}
\label{alg:active_pose_tracking}
\begin{algorithmic}[1]
\Require
  Object pose in camera frame ${}^cT_o$,
  initial end-effector pose ${}^bT_{e,1}$,
  camera pose in end-effector frame ${}^eT_c$, history length $H$, prediction horizon $K$, executed tail length $k_h$, total steps $N$.
\Ensure
  executed poses $\{{}^bT_{e,t}\}_{t=1}^{N}$

\State $t \gets H$
\While{$t \le N$}
  \State Compute object pose in base frame:
    ${}^bT_o$ (Eq.~\eqref{eq:base_object_pose})
  \State Collect history window:
    $O_t$
  \State Sample a future end-effector pose sequence by denoising:
    \[
      \{{}^b\widehat T_e^{\,t+1:t+K}\} \leftarrow \mathrm{DP}(O_t)
    \]
    \Comment{DP denotes the diffusion policy, see Eqs.~\eqref{eq:reverse_step}--\eqref{eq:loss}}
  \State Select the \textbf{last $k_h$ poses} of the predicted horizon  $K$:
    \[
      \mathcal{T}_t \;\gets\; \{{}^b\widehat T_e^{\,t+K-k_h+1:t+K}\}
    \]
  \State Time-parameterize and interpolate from current ${}^bT_e$ to $\mathcal{T}_t$, then execute the resulting continuous trajectory via servo control.
  \State Update ${}^bT_e$ to the last executed pose:
    ${}^bT_e \leftarrow {}^b\widehat T_e^{\,t+K}$
  \State $t \leftarrow t + k_h$
\EndWhile
\end{algorithmic}
\end{algorithm}

After obtaining the final disambiguated pose ${}^{c}\!T_{o}$ from active pose estimation, we formulate active tracking as a \textit{conditional diffusion} problem to maintain continuous visibility (see Fig.~\ref{fig:active_pose_tracking}). Because the camera is rigidly mounted on the sensing end-effector with a fixed extrinsic ${}^eT_c$,
a predicted end-effector trajectory $\{{}^bT_{e,t}\}$ uniquely induces a camera trajectory $\{{}^bT_{c,t}\}$ via
${}^bT_c = {}^bT_e \cdot {}^eT_c$.
For clarity, we learn and execute end-effector trajectories, which are equivalent to camera trajectories under this rigid mounting.

At each time step $t$, the pose estimator provides the object pose in the camera frame ${}^cT_o$.
We convert it to the robot base frame using the current end-effector pose ${}^bT_e$ and the fixed camera extrinsic ${}^eT_c$:
\begin{equation}
{}^bT_o = {}^bT_c \cdot {}^cT_o
      = {}^bT_e \cdot {}^eT_c \cdot {}^cT_o .
\label{eq:base_object_pose}
\end{equation}
We then build the observation
$O_t=\bigl\{{}^bT_{o}^{\,t-H+1:t},\;{}^bT_{e}^{\,t-H+1:t}\bigr\}$,
containing the past $H$ frames of (i) the object pose in the robot base frame ${}^bT_o$ and (ii) the end-effector pose in the base frame ${}^bT_e$. We condition the policy on a short history of poses rather than a single-frame estimate, which provides motion context and improves tracking robustness.

Specifically, starting from Gaussian noise $A_t^{K_d}$, the denoising network $\varepsilon_\theta$ performs $K_d$ reverse-diffusion steps to transform $A_t^{K_d}$ into a noise-free action $A_t^{0}$:
\begin{align}
  A_t^{k-1}
    &= \alpha_k\Bigl(A_t^k - \gamma_k\,\varepsilon_\theta(O_t, A_t^k, k)\Bigr)
      + \sigma_k\,\mathcal{N}(0,I),
  \label{eq:reverse_step}\\
  \mathcal{L}
    &= \mathbb{E}\Bigl[\bigl\|\epsilon^k - \varepsilon_\theta\bigl(O_t,\;\hat\alpha_k A_t^0 + \hat\beta_k \epsilon^k,\;k\bigr)\bigr\|^2\Bigr],
  \label{eq:loss}
\end{align}
where $\alpha_k,\gamma_k,\sigma_k$ are parameters of the noise schedule at step $k$, $\epsilon^k$ is the injected noise, and $\hat\alpha_k,\hat\beta_k$ are the one-step forward diffusion coefficients~\cite{song2022denoisingdiffusionimplicitmodels}. Mean squared error (MSE) loss is employed as the objective function to supervise noise prediction (Eq.~\eqref{eq:loss}). 
The rotation is represented as R6D \cite{9578682}, a continuous 6D representation for $SO(3)$ that avoids discontinuities and is empirically stable for generative modeling.
At runtime, the diffusion policy outputs an absolute horizon $\{{}^b\widehat T_e^{\,t+1:t+K}\}$.
Instead of executing a prefix, we execute the \emph{last} $k_h$ poses $\{{}^b\widehat T_e^{\,t+K-k_h+1:t+K}\}$, biasing the camera toward the terminal viewpoint that best preserves future visibility.
We interpolate from the current ${}^bT_e$ to this segment for smooth, feasible motion and repeat in a receding-horizon loop.

\section{EXPERIMENTS}
This section describes the experimental setups, implementation details, evaluation protocol, and quantitative results for active pose estimation and active pose tracking, followed by ablations, runtime analysis, and an engineering case study on peg-in-hole assembly.

\subsection{Experimental Setups}

\begin{figure}[htbp]
  \centering
  \includegraphics[scale=0.277]{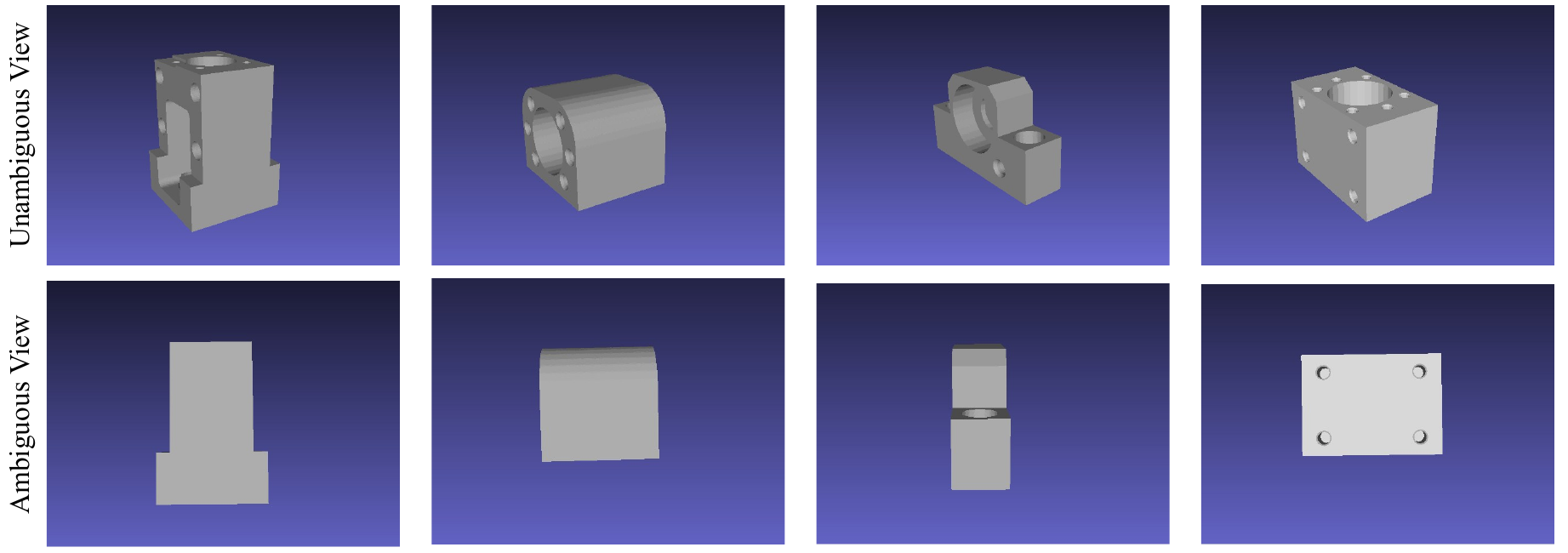}
  \caption{CAD models of the experimental objects (Obj.\ 1--3 from MP6D; Obj.\ 4 for assembly) and an example ambiguous view.}
  \label{Experimental Objects}
\end{figure}

\textbf{Hardware/Software and Objects.}
For \textbf{active pose estimation}, we evaluate in both simulation and the real world.
Simulation is implemented in PyBullet \cite{panerati2021learningflygym}, which provides controlled rendering for NBV selection.
For \textbf{active pose tracking}, we exclusively perform real-robot evaluations to capture realistic occlusions, dynamics, and sensing noise.
All real-world experiments use two Franka Emika Panda 7-DOF arms: a \emph{sensing arm} carries a wrist-mounted Intel RealSense D435 RGB-D camera (640$\times$480 at 30\,Hz), and a \emph{manipulation arm} is equipped with a parallel-jaw gripper (see Fig.~\ref{workspace}).
As shown in Fig.~\ref{Experimental Objects}, we select three objects from MP6D \cite{9722997} for evaluation and an additional object for the peg-in-hole assembly task.
All four objects exhibit strong viewpoint-induced ambiguities due to symmetry and textureless geometry.

\textbf{Implementation details.}
\emph{Active pose estimation:}
We call the ChatGPT4o API as our VLM.
The VLM operates on a cropped image (resized to 224$\times$224) paired with a fixed, per-object geometry-aware prompt, while FoundationPose uses the full RGB-D observation. VLM inference is run deterministically: we set $\texttt{temperature}=0$ (and $\texttt{top\_p}=1.0$) and compute the ambiguity probability from the returned \{Yes, No\} log-probabilities 
(Eq.~\eqref{eq:p}); all other decoding options are fixed across experiments.
Offline prompting uses $K{=}48$ rendered views, from which we select $K_u{=}K_a{=}2$ exemplars; the ambiguity threshold is fixed to $\tau{=}0.5$.
When ambiguity is triggered, we render $M{=}12$ feasible candidate viewpoints and score them with fusion weight $\lambda{=}0.6$. We use fixed $L{=}3$ across all objects and scenarios.

\emph{Prompt template:}
For each object, the prompt consists of (i) a short instruction asking whether the pose is ambiguous and constraining the answer to \{\texttt{Yes}, \texttt{No}\}, (ii) $K_u$ low-entropy exemplar renders labeled \texttt{No} and $K_a$ high-entropy exemplar renders labeled \texttt{Yes}, and (iii) the query image.

\emph{Active pose tracking:}
We collect 10 expert demonstrations per task, each consisting of $\sim$200 frames of object poses ${}^bT_o$ and end-effector poses ${}^bT_e$ at 10\,Hz, via teleoperation that keeps the target within the camera frustum.
We train the diffusion policy with AdamW (batch size 40, learning rate $3\!\times\!10^{-4}$) for 2{,}000 epochs.
We use fixed $H{=}5$, $K{=}10$, $k_h{=}5$ and $N{=}200$ across all objects and scenarios.
All training and inference run on a single NVIDIA RTX 4090 GPU.

\textbf{Estimation scenarios.}
We evaluate active pose estimation under two conditions:
\begin{enumerate}
  \item \emph{Random Placement:} each object is placed at a random position and orientation on the workspace.
  \item \emph{High-Entropy Placement:} each object is deliberately positioned at a precomputed high-entropy (ambiguous) viewpoint.
\end{enumerate}

\textbf{Estimation baselines.}
\begin{itemize}
  \item \textbf{Fixed-View:} run FoundationPose once at the initial camera pose (no active viewpoint selection).
  \item \textbf{Random-NBV:} upon ambiguity detection, randomly select one of the $M$ feasible candidate viewpoints for a second FoundationPose estimate.
  \item \textbf{Entropy-NBV:} upon ambiguity detection, select NBV by minimizing the normalized hypothesis entropy $\bar H_j$ over the $M$ feasible candidates (no VLM).
  \item \textbf{VLM-NBV:} upon ambiguity detection, select NBV by minimizing VLM ambiguity $p_{\mathrm{amb},j}$ over the $M$ feasible candidates (no FoundationPose).
\end{itemize}

\textbf{Estimation metric.}
We report \emph{success rate} (SR), defined as the percentage of trials whose final estimated 6D pose has translation error below 5\,mm and rotation error below $5^\circ$.
Each method is tested for 100 trials per scenario.

\textbf{Tracking scenarios.}
We evaluate all trackers under four challenging conditions:
\begin{enumerate}
  \item \emph{Long-range linear motion:} the object moves along a straight trajectory from point A to point B, where point B is randomized.
  \item \emph{Circular rotational motion:} the object rotates around a fixed point on the table, inducing large viewpoint changes.
  \item \emph{Temporary occlusion:} the object is briefly occluded (up to 1\,s), requiring recovery upon reappearance.
  \item \emph{Random spatial motion:} the object undergoes unpredictable translational and rotational motions in 3D, including sharp turns and speed variations.
\end{enumerate}

\textbf{Tracking baselines.}
\begin{itemize}
  \item \textbf{Pose-Servo:} a classical pose-based visual servoing baseline. The wrist-mounted camera provides the current 6D object pose; a proportional controller maps pose error to end-effector velocity commands.
  \item \textbf{World-Camera:} a fixed, calibrated overhead camera continuously tracks the object's 6D pose via FoundationPose.
\end{itemize}

\textbf{Tracking metric.}
We report \emph{success rate} (SR) per trial: the fraction of runs that complete the scenario without any \emph{pose-loss} event.
We declare pose-loss if (i) FoundationPose returns no valid pose for more than 10 consecutive frames, or (ii) the object exits the camera FOV.
Each method is evaluated in 50 independent trials per scenario.

\begin{figure}[htbp]
  \centering
  \includegraphics[scale=0.49]{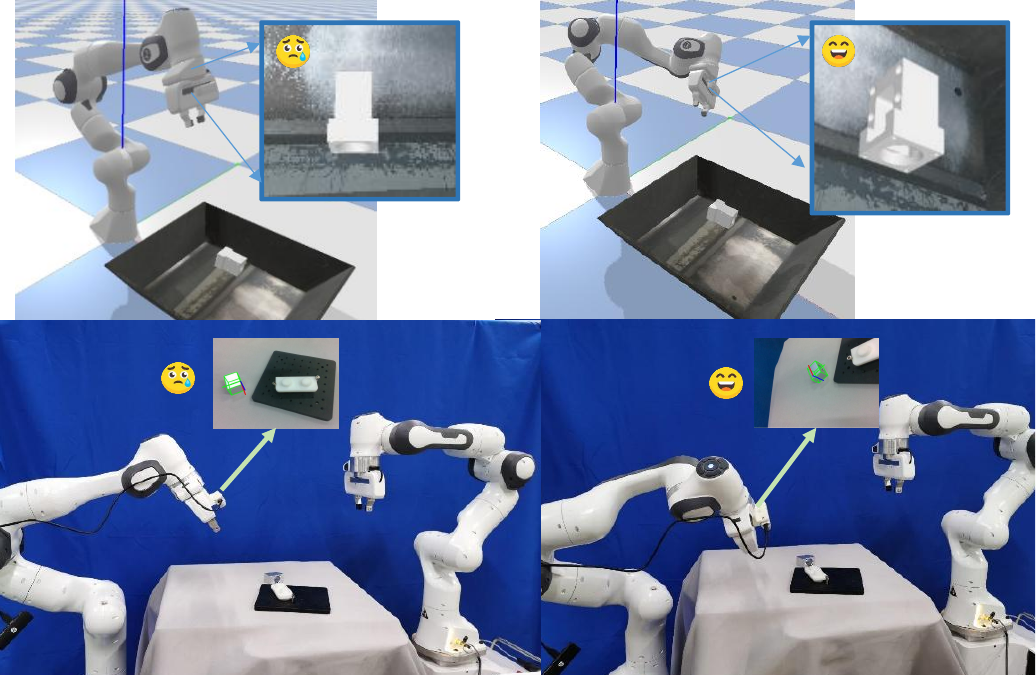}
  \caption{\textbf{Example of active pose estimation.} An ambiguous initial view triggers NBV selection; moving to the selected viewpoint yields an unambiguous 6D pose estimate, shown in simulation (top) and real-robot trials (bottom).}
  \label{example_of_estimation}
\end{figure}

\begin{figure*}[htbp]
  \centering
  \includegraphics[width=\textwidth]{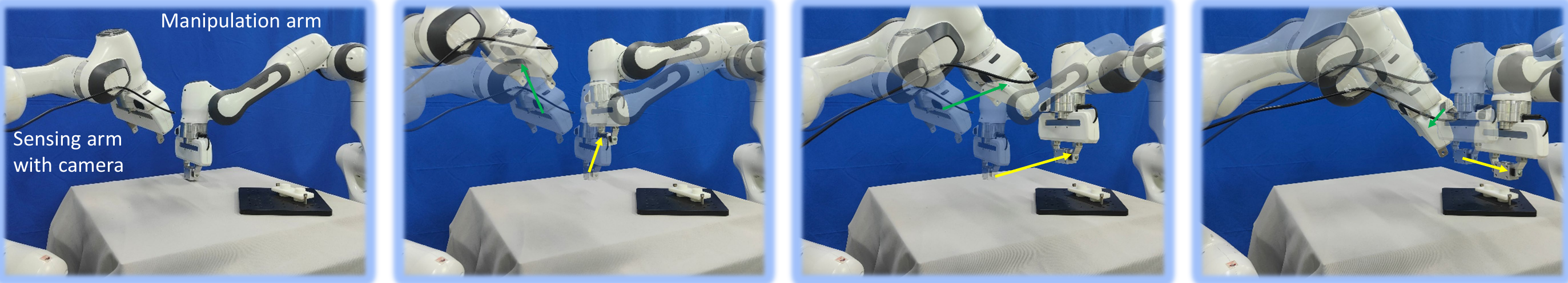}
  \caption{\textbf{Example of active pose tracking.} While the manipulation arm manipulates the object, the sensing arm with a wrist-mounted RGB-D camera actively follows to keep the target in view.}
  \label{experiment_pipeline}
\end{figure*}

% --- tables unchanged ---
% =========================

\begin{table*}[t]
  \renewcommand{\arraystretch}{1.15}
  \centering
  \caption{\textbf{Active pose estimation in simulation and on real robots.}}
  \label{tab:estimation_sim_real}
  \setlength{\tabcolsep}{3.6pt}
  \resizebox{\textwidth}{!}{%
  \begin{tabular}{l|ccccc|ccccc|ccccc|ccccc}
    \hline
    \multirow{2}{*}{Method}
    & \multicolumn{10}{c|}{Simulation}
    & \multicolumn{10}{c}{Real Robots} \\
    \cline{2-11}\cline{12-21}
    & \multicolumn{5}{c|}{Random Placement}
    & \multicolumn{5}{c|}{High-Entropy Placement}
    & \multicolumn{5}{c|}{Random Placement}
    & \multicolumn{5}{c}{High-Entropy Placement} \\
    \cline{2-6}\cline{7-11}\cline{12-16}\cline{17-21}
    & Obj 1 & Obj 2 & Obj 3 & Obj 4 & SR
    & Obj 1 & Obj 2 & Obj 3 & Obj 4 & SR
    & Obj 1 & Obj 2 & Obj 3 & Obj 4 & SR
    & Obj 1 & Obj 2 & Obj 3 & Obj 4 & SR \\
    \hline
    Fixed-View
      & 70.0\% & 50.0\% & 50.0\% & 70.0\% & 60.0\%
      & 40.0\% & 10.0\% & 20.0\% & 10.0\% & 20.0\%
      & 68.0\% & 47.0\% & 42.0\% & 54.0\% & 52.8\%
      & 26.0\% & 21.0\% & 31.0\% & 9.0\%  & 21.8\% \\
    Random-NBV
      & 73.0\% & 51.0\% & 66.0\% & 70.0\% & 65.0\%
      & 41.0\% & 21.0\% & 34.0\% & 17.0\% & 28.3\%
      & 71.0\% & 52.0\% & 43.0\% & 64.0\% & 57.5\%
      & 55.0\% & 41.0\% & 67.0\% & 28.0\% & 47.8\% \\
    Entropy-NBV
      & 65.0\% & 57.0\% & 61.0\% & 74.0\% & 64.3\%
      & 52.0\% & 38.0\% & 41.0\% & 37.0\% & 42.0\%
      & 73.0\% & 51.0\% & 50.0\% & 57.0\% & 57.8\%
      & 60.0\% & 46.0\% & 63.0\% & 26.0\% & 48.8\% \\
    VLM-NBV
      & 67.0\% & 53.0\% & 25.0\% & 19.0\% & 41.0\%
      & 46.0\% & 34.0\% & 27.0\% & 32.0\% & 34.8\%
      & 37.0\% & 29.0\% & 23.0\% & 41.0\% & 32.5\%
      & 31.0\% & 37.0\% & 51.0\% & 10.0\% & 32.3\% \\
    \rowcolor{gray!20} Ours
      & \textbf{100.0\%} & \textbf{100.0\%} & \textbf{100.0\%} & \textbf{90.0\%} & \textbf{97.5\%}
      & \textbf{100.0\%} & \textbf{90.0\%}  & \textbf{100.0\%} & \textbf{90.0\%} & \textbf{95.0\%}
      & \textbf{90.0\%}  & \textbf{90.0\%}  & \textbf{100.0\%} & \textbf{90.0\%} & \textbf{92.5\%}
      & \textbf{100.0\%} & \textbf{90.0\%}  & \textbf{100.0\%} & \textbf{90.0\%} & \textbf{95.0\%} \\
    \hline
  \end{tabular}%
  }
\end{table*}
% =========================
% Table 3: Tracking (Real)
% =========================
\begin{table*}[t]
  \renewcommand{\arraystretch}{1.25}
  \centering
  \caption{\textbf{Active pose tracking on real robots.}}
  \label{a_track_real}
  \setlength{\tabcolsep}{3.6pt}
  \resizebox{\textwidth}{!}{%
  \begin{tabular}{l|cccc|c|cccc|c|cccc|c|cccc|c}
    \hline
    Method
    & \multicolumn{5}{c|}{Long-range Linear Motion}
    & \multicolumn{5}{c|}{Circular Rotational Motion}
    & \multicolumn{5}{c|}{Temporary Occlusion}
    & \multicolumn{5}{c}{Random Spatial Motion} \\
    \cline{2-6}\cline{7-11}\cline{12-16}\cline{17-21}
    & Obj 1 & Obj 2 & Obj 3 & Obj 4 & SR
    & Obj 1 & Obj 2 & Obj 3 & Obj 4 & SR
    & Obj 1 & Obj 2 & Obj 3 & Obj 4 & SR
    & Obj 1 & Obj 2 & Obj 3 & Obj 4 & SR \\
    \hline
    Pose-Servo
      & 65.0\% & 70.0\% & 60.0\% & 50.0\% & 61.3\%
      & 0.0\%  & 0.0\%  & 0.0\%  & 0.0\%  & 0.0\%
      & 0.0\%  & 0.0\%  & 5.0\%  & 0.0\%  & 1.3\%
      & 50.0\% & 45.0\% & 50.0\% & 55.0\% & 50.0\% \\
    World-Camera
      & 50.0\% & 45.0\% & 65.0\% & 35.0\% & 48.8\%
      & 65.0\% & 60.0\% & 75.0\% & 50.0\% & 62.5\%
      & 35.0\% & 15.0\% & 15.0\% & 5.0\%  & 17.5\%
      & 70.0\% & 55.0\% & 65.0\% & 45.0\% & 58.8\% \\
    \rowcolor{gray!20} Ours
      & \textbf{85.0\%} & \textbf{90.0\%} & \textbf{80.0\%} & \textbf{95.0\%} & \textbf{87.5\%}
      & \textbf{80.0\%} & \textbf{90.0\%} & \textbf{95.0\%} & \textbf{100.0\%} & \textbf{91.3\%}
      & \textbf{50.0\%} & \textbf{45.0\%} & \textbf{65.0\%} & \textbf{50.0\%} & \textbf{52.5\%}
      & \textbf{70.0\%} & \textbf{75.0\%} & \textbf{65.0\%} & \textbf{80.0\%} & \textbf{72.5\%} \\
    \hline
  \end{tabular}%
  }
\end{table*}

\begin{table*}[t]
  \renewcommand{\arraystretch}{1.15}
  \centering
  \caption{\textbf{Ablation of active pose estimation.}}
  \label{tab:ablation_est_new}
  \setlength{\tabcolsep}{4.6pt}
  \begin{tabular}{l|cc|cc|cc|cc}
    \hline
    \multirow{2}{*}{Method} &
    \multicolumn{2}{c|}{Sim / Random} &
    \multicolumn{2}{c|}{Sim / High-Entropy} &
    \multicolumn{2}{c|}{Real / Random} &
    \multicolumn{2}{c}{Real / High-Entropy} \\
    \cline{2-9}
    & SR  & \#NBV 
    & SR  & \#NBV 
    & SR  & \#NBV 
    & SR  & \#NBV  \\
    \hline
    \rowcolor{gray!20} Ours (entropy-ranked exemplars, $\lambda{=}0.6$) &
    \textbf{97.5\%} & \textbf{0.60} &
    \textbf{95.0\%} & \textbf{0.90} &
    \textbf{92.5\%} & \textbf{0.60} &
    \textbf{95.0\%} & \textbf{1.00} \\
    \hline
    \multicolumn{9}{l}{\textbf{Prompt grounding (exemplars)}}\\
    \hline
    Text-only prompt (no exemplars) &
    73.0\% & 1.60 &
    61.0\% & 1.30 &
    70.5\% & 1.65 &
    53.0\% & 2.00 \\
    Random exemplars (same count as $K_u{+}K_a$) &
    77.5\% & 1.10 &
    68.0\% & 1.00 &
    78.5\% & 0.90 &
    70.5\% & 1.10 \\
    \hline

    \multicolumn{9}{l}{\textbf{Fusion sensitivity (sweep $\lambda$)}}\\
    \hline
    $\lambda{=}0$ (VLM-only) &
    41.0\% & 2.90 &
    34.8\% & 3.00 &
    32.5\% & 2.80 &
    32.3\% & 3.00 \\
    $\lambda{=}0.3$ &
    47.4\% & 2.70 &
    36.8\% & 2.60 &
    31.6\% & 2.60 &
    25.4\% & 2.50 \\
    $\lambda{=}0.9$ &
    61.7\% & 1.30 &
    46.0\% & 1.40 &
    65.5\% & 1.40 &
    50.0\% & 1.70 \\
    $\lambda{=}1$ (entropy-only) &
    64.3\% & 1.40 &
    42.0\% & 1.60 &
    57.8\% & 1.50 &
    48.8\% & 1.30 \\
    \hline
  \end{tabular}
\end{table*}

\subsection{Experimental Results}
\textbf{Active Pose Estimation Results.}
We evaluate two placement conditions: \textit{Random Placement} and \textit{High-Entropy Placement} (deliberately ambiguous starts).
Unless otherwise stated, SR is averaged over four objects (Obj.~1--4).

\emph{Simulation.}
As reported in Table~\ref{tab:estimation_sim_real}, Fixed-View degrades sharply from Random Placement (60.0\%) to High-Entropy Placement (20.0\%), highlighting strong viewpoint-induced ambiguity.
Random-NBV (28.3\%) improves over Fixed-View (20.0\%), confirming that acquiring an additional view can help, but remains unreliable without guided selection.
Entropy-NBV performs competitively under Random Placement (64.3\%) but drops under High-Entropy Placement (42.0\%), suggesting that hypothesis entropy alone can be an imperfect proxy for selecting a truly disambiguating view.
VLM-NBV performs worst among active baselines (41.0\% / 34.8\%), indicating that VLM ambiguity alone is insufficient for reliable NBV ranking over rendered candidates.
In contrast, ActivePose (97.5\% / 95.0\%) achieves consistently high SR in both conditions, demonstrating robust disambiguation (Table~\ref{tab:estimation_sim_real}).

\emph{Real world.}
Table~\ref{tab:estimation_sim_real} shows a similar trend: Fixed-View (21.8\%) and Random-NBV (47.8\%) drop substantially under High-Entropy Placement.
Entropy-NBV yields moderate gains over Random-NBV in the High-Entropy setting (48.8\% vs.\ 47.8\%), while VLM-NBV remains unreliable in the real world (32.5\% / 32.3\%), consistent with the simulation results.
In contrast, ActivePose achieves the best performance in both placements (92.5\% on Random and 95.0\% on High-Entropy), indicating robust disambiguation.
The lower SR in real-world trials compared to simulation is likely due to real sensing imperfections, including sensor noise and the synthetic-to-real gap between rendered and captured observations. Figure~\ref{example_of_estimation} shows an example of active pose estimation, where an ambiguous view triggers NBV selection and a second view resolves the ambiguity.

\textbf{Active Pose Tracking Results.}
Table~\ref{a_track_real} reports SR 1n 50 trials per scenario.
Across all four motion scenarios, our tracker consistently outperforms Pose-Servo and World-Camera by large margins.
Pose-Servo frequently fails due to reachability limitations under large viewpoint changes, while World-Camera fails when the object exits its fixed FOV.
Unlike pose-based servoing, which enforces strict instantaneous pose-error tracking and can drive the arm into infeasible motions, our learned policy generates smooth, anticipatory camera motions that prioritize maintaining visibility and reacquiring the target after occlusion. Figure~\ref{experiment_pipeline} illustrates an example of active pose tracking in our dual-arm setup.

\subsection{Ablation Studies}
To isolate the specific value of using a VLM, to examine whether the VLM ambiguity score complements the FoundationPose entropy signal, and to assess the choice of the fusion weight $\lambda$, we conduct ablations on the active pose estimation module.

We ablate active pose estimation on the same four objects under both placement conditions, using the same SR metric (5\,mm / 5$^\circ$) and default settings ($K{=}48$, $K_u{=}K_a{=}2$, $M{=}12$, $\tau{=}0.5$, $L{=}3$).
In addition to SR, we report \#NBV: the average number of executed NBV actions (additional viewpoints beyond the initial view) per trial. A lower \#NBV is better.

\textbf{Prompt grounding.}
Table~\ref{tab:ablation_est_new} shows that entropy-ranked exemplars substantially outperform a text-only prompt and random exemplars, improving both SR and \#NBV. Random exemplars keep the same exemplar count and labels as the full method, but randomly sample the low-/high-entropy exemplars from their respective pools instead of taking the entropy extremes.
Notably, text-only prompting frequently exhausts the NBV budget (\#NBV $\approx 1.3$--$2.0$) while still yielding low SR, indicating that exemplars are critical for calibrating the VLM to \emph{geometric} ambiguity.

\textbf{Fusion sensitivity.}
Sweeping $\lambda$ confirms that intermediate fusion ($\lambda{=}0.6$) provides the best SR--\#NBV trade-off.
At $\lambda{=}0$ (VLM-only), the policy often hits the budget limit (\#NBV $\approx 2.5$--$3.0$) yet remains inaccurate, showing that VLM ambiguity alone is insufficient for reliable NBV ranking.
At $\lambda{=}1$ (entropy-only), \#NBV decreases but SR remains substantially below the fused setting, suggesting that hypothesis entropy alone can be misled by rendering-to-real discrepancies and partial occlusions.
Overall, the results support that VLM ambiguity and hypothesis entropy provide complementary cues for selecting disambiguating viewpoints.

\begin{table}[t] 
\centering 
\caption{\textbf{Runtime breakdown of active pose estimation.}}
\label{tab:runtime} 
\setlength{\tabcolsep}{4pt} \renewcommand{\arraystretch}{1.1} \begin{tabular}{l|c} 
\hline Component & Time (ms) \\ 
\hline FoundationPose (single view, per image) & 60 \\ 
Render $M{=}12$ imagined views (total) & 80 \\ 
VLM query (per call, mean $\pm$ std) & $600 \pm 200$ 
\\ 
NBV evaluation (12 VLM calls + scoring) & 7,500 \\ 
End-to-end NBV cycle (worst case, incl.\ motion) & 11,000 \\ 
\hline 
\end{tabular} 
\end{table}

% =========================
% Table 5: Assembly
% =========================
\begin{table}[t]
  \renewcommand{\arraystretch}{1.15}
  \centering
  \caption{\textbf{Peg-in-hole assembly.}}
  \label{assembly-sr}
  \setlength{\tabcolsep}{6pt}
  \begin{tabular}{l|c}
    \hline
    Method & SR \\
    \hline
    Fixed-View + World-Camera & 40\% \\
    Fixed-View + Pose-Servo & 50\% \\
    Random-NBV + Pose-Servo & 70\% \\
    \rowcolor{gray!20} Ours & 90\% \\
    \hline
  \end{tabular}
\end{table}

\subsection{Runtime Analysis}

Querying an external VLM API introduces response delays, which can be a practical limitation for real-time robotics applications.
Table~\ref{tab:runtime} reports a runtime breakdown of active pose estimation on a workstation with an NVIDIA RTX 4090 GPU.
In the worst case (ambiguity detected and all $M{=}12$ candidates evaluated), one NBV cycle takes $\sim$11{,}000\,ms including robot motion.

In our pipeline, this latency has limited impact on overall task completion for three reasons.
First, active disambiguation is triggered only when $p_{\mathrm{amb}}>\tau$.
Second, disambiguation is performed only at grasp initialization or after pose-loss recovery, rather than inside the high-frequency tracking loop.
Third, the compute time within an NBV cycle is dominated by $M$ sequential VLM queries for candidate ranking, which increases the \emph{time-to-disambiguate} but does not affect per-step tracking control once a reliable pose is obtained.

Overall, while VLM queries remain the main bottleneck for faster disambiguation, they are not on the critical path of per-step tracking control and thus do not materially hinder manipulation in our scenarios.

\subsection{Engineering Case Study: Peg-in-Hole Assembly}

Beyond isolated estimation and tracking, we evaluate a peg-in-hole assembly task as an \emph{engineering case study} to demonstrate the practical utility of ActivePose in a closed-loop manipulation pipeline.
In each run, the robot grasps the object from the tabletop and inserts it into a fixed socket whose position is randomized.
We use the assembly policy from \cite{sun2025exploringposeguidedimitationlearning}, which takes as input the relative SE(3) pose between the grasped object and the socket and outputs short-horizon end-effector motions to complete the assembly task.
ActivePose provides this relative pose online by combining (i) active disambiguation at grasp initialization and (ii) active tracking during insertion to maintain visibility under motion and occlusions.
This setting jointly tests reliable initialization (resolving grasp-time ambiguity) and sustained visibility during insertion.

\textbf{Baselines and metric.}
We compare against three baselines: \textbf{Fixed-View + World-Camera} (single estimate from a fixed world camera), \textbf{Fixed-View + Pose-Servo} (single estimate at grasp followed by pose-based visual servoing during insertion), and \textbf{Random-NBV + Pose-Servo} (one random feasible NBV at grasp for a second estimate, then pose-servo during insertion).
We report success rate (SR) as the fraction of trials that complete the insertion (20 trials per method).

\textbf{Results and discussion.}
Table~\ref{assembly-sr} shows that ActivePose achieves the best performance (90\%).
Fixed-View + World-Camera often fails when the randomized socket falls outside the fixed FOV (40\%), while Fixed-View + Pose-Servo is limited by reachability and pose-loss during insertion (50\%).
Random-NBV + Pose-Servo reduces grasp-time ambiguity and improves SR to 70\%, but still suffers from pose-loss under a fixed-view servo controller.
By actively disambiguating at grasp and tracking during insertion, ActivePose maintains visibility and provides reliable relative poses for the downstream insertion.

\section{Conclusion}
In this work, we propose \textit{ActivePose}, a closed-loop framework that combines zero-shot active pose estimation with active camera tracking for robotic manipulation.
Our method performs pose estimation by grounding a VLM with entropy-ranked CAD renderings and selecting disambiguating NBVs, and then performs tracking to maintain visibility during manipulation using a diffusion-policy tracker to prevent pose-loss under motion and occlusions.
Experiments in simulation and on real dual-arm hardware show consistent improvements over baselines, and an industrial peg-in-hole assembly case study demonstrates practical benefits in a closed-loop manipulation pipeline.

%Future work will focus on reducing reliance on external VLM APIs by developing lightweight, locally deployable ambiguity predictors and speeding up NBV evaluation. 

\bibliographystyle{ieeetr}
\bibliography{my}   

@misc{wolf2025diffusionmodelsroboticmanipulation,
      title={Diffusion Models for Robotic Manipulation: A Survey}, 
      author={Rosa Wolf and Yitian Shi and Sheng Liu and Rania Rayyes},
      year={2025},
      eprint={2504.08438},
      archivePrefix={arXiv},
      primaryClass={cs.RO},
      url={https://arxiv.org/abs/2504.08438}, 
}

@inproceedings{peng2019pvnet,
	title={Pvnet: Pixel-wise voting network for 6dof pose estimation},
	author={Peng, Sida and Liu, Yuan and Huang, Qixing and Zhou, Xiaowei and Bao, Hujun},
	booktitle={Proceedings of the IEEE/CVF Conference on Computer Vision and Pattern Recognition},
	pages={4561--4570},
	year={2019}
}

@inproceedings{fang2020graspnet,
	title={Graspnet-1billion: A large-scale benchmark for general object grasping},
	author={Fang, Hao-Shu and Wang, Chenxi and Gou, Minghao and Lu, Cewu},
	booktitle={Proceedings of the IEEE/CVF conference on computer vision and pattern recognition},
	pages={11444--11453},
	year={2020}
}

@inproceedings{Mizuno_2024,
   title={Object Pose Estimation by Camera Arm Control Based on the Next Viewpoint Estimation},
   url={http://dx.doi.org/10.1109/IROS58592.2024.10801633},
   DOI={10.1109/iros58592.2024.10801633},
   booktitle={2024 IEEE/RSJ International Conference on Intelligent Robots and Systems (IROS)},
   publisher={IEEE},
   author={Mizuno, Tomoki and Yabashi, Kazuya and Tasaki, Tsuyoshi},
   year={2024},
   month=oct, pages={9482–9487} }

@article{choi2018learning,
	title={Learning object grasping for soft robot hands},
	author={Choi, Changhyun and Schwarting, Wilko and DelPreto, Joseph and Rus, Daniela},
	journal={IEEE Robotics and Automation Letters},
	volume={3},
	number={3},
	pages={2370--2377},
	year={2018},
	publisher={IEEE}
}

@article{zhang20216dof,
	title={A 6DoF Pose Estimation Dataset and Network for Multiple Parametric Shapes in Stacked Scenarios},
	author={Zhang, Xinyu and Lv, Weijie and Zeng, Long},
	journal={Machines},
	volume={9},
	number={12},
	pages={321},
	year={2021},
	publisher={Multidisciplinary Digital Publishing Institute}
}

@article{malik2019advances,
	title={Advances in machine vision for flexible feeding of assembly parts},
	author={Malik, Ali Ahmad and Andersen, Martin Vejling and Bilberg, Arne},
	journal={Procedia Manufacturing},
	volume={38},
	pages={1228--1235},
	year={2019},
	publisher={Elsevier}
}

@inproceedings{zakharov2019dpod,
	title={Dpod: 6d pose object detector and refiner},
	author={Zakharov, Sergey and Shugurov, Ivan and Ilic, Slobodan},
	booktitle={Proceedings of the IEEE/CVF international conference on computer vision},
	pages={1941--1950},
	year={2019}
}

@ARTICLE{9565319,
	author={Shugurov, Ivan and Zakharov, Sergey and Ilic, Slobodan},
	journal={IEEE Transactions on Pattern Analysis and Machine Intelligence}, 
	title={DPODv2: Dense Correspondence-Based 6 DoF Pose Estimation}, 
	year={2022},
	volume={44},
	number={11},
	pages={7417-7435},
	doi={10.1109/TPAMI.2021.3118833}}

@inproceedings{park2019pix2pose,
	title={Pix2pose: Pixel-wise coordinate regression of objects for 6d pose estimation},
	author={Park, Kiru and Patten, Timothy and Vincze, Markus},
	booktitle={Proceedings of the IEEE/CVF International Conference on Computer Vision},
	pages={7668--7677},
	year={2019}
}

@inproceedings{li2019cdpn,
	title={Cdpn: Coordinates-based disentangled pose network for real-time rgb-based 6-dof object pose estimation},
	author={Li, Zhigang and Wang, Gu and Ji, Xiangyang},
	booktitle={Proceedings of the IEEE/CVF International Conference on Computer Vision},
	pages={7678--7687},
	year={2019}
}

@article{cheng2021real,
	title={Real-time and efficient 6-D pose estimation from a single RGB image},
	author={Cheng, Jun and Liu, Penglei and Zhang, Qieshi and Ma, Hui and Wang, Fei and Zhang, Jin},
	journal={IEEE Transactions on Instrumentation and Measurement},
	volume={70},
	pages={1--14},
	year={2021},
	publisher={IEEE}
}

@inproceedings{wang2021gdr,
	title={Gdr-net: Geometry-guided direct regression network for monocular 6d object pose estimation},
	author={Wang, Gu and Manhardt, Fabian and Tombari, Federico and Ji, Xiangyang},
	booktitle={Proceedings of the IEEE/CVF Conference on Computer Vision and Pattern Recognition},
	pages={16611--16621},
	year={2021}
}

@article{zhou20226,
	title={6-D Object Pose Estimation Using Multiscale Point Cloud Transformer},
	author={Zhou, Guangliang and Wang, Deming and Yan, Yi and Liu, Chengju and Chen, Qijun},
	journal={IEEE Transactions on Instrumentation and Measurement},
	volume={72},
	pages={1--11},
	year={2022},
	publisher={IEEE}
}

@inproceedings{wen2024foundationpose,
	title={Foundationpose: Unified 6d pose estimation and tracking of novel objects},
	author={Wen, Bowen and Yang, Wei and Kautz, Jan and Birchfield, Stan},
	booktitle={Proceedings of the IEEE/CVF Conference on Computer Vision and Pattern Recognition},
	pages={17868--17879},
	year={2024}
}

@inproceedings{lin2024sam,
	title={Sam-6d: Segment anything model meets zero-shot 6d object pose estimation},
	author={Lin, Jiehong and Liu, Lihua and Lu, Dekun and Jia, Kui},
	booktitle={Proceedings of the IEEE/CVF Conference on Computer Vision and Pattern Recognition},
	pages={27906--27916},
	year={2024}
}

@inproceedings{nguyen2024gigapose,
	title={Gigapose: Fast and robust novel object pose estimation via one correspondence},
	author={Nguyen, Van Nguyen and Groueix, Thibault and Salzmann, Mathieu and Lepetit, Vincent},
	booktitle={Proceedings of the IEEE/CVF Conference on Computer Vision and Pattern Recognition},
	pages={9903--9913},
	year={2024}
}

@inproceedings{hofer2023hyperposepdf,
	title={Hyperposepdf-hypernetworks predicting the probability distribution on so (3)},
	author={H{\"o}fer, Timon and Kiefer, Benjamin and Messmer, Martin and Zell, Andreas},
	booktitle={Proceedings of the IEEE/CVF Winter Conference on Applications of Computer Vision},
	pages={2369--2379},
	year={2023}
}

@article{murphy2021implicit,
	title={Implicit-pdf: Non-parametric representation of probability distributions on the rotation manifold},
	author={Murphy, Kieran and Esteves, Carlos and Jampani, Varun and Ramalingam, Srikumar and Makadia, Ameesh},
	journal={arXiv preprint arXiv:2106.05965},
	year={2021}
}

@inproceedings{hodavn2020bop,
	title={BOP challenge 2020 on 6D object localization},
	author={Hoda{\v{n}}, Tom{\'a}{\v{s}} and Sundermeyer, Martin and Drost, Bertram and Labb{\'e}, Yann and Brachmann, Eric and Michel, Frank and Rother, Carsten and Matas, Ji{\v{r}}{\'\i}},
	booktitle={Computer Vision--ECCV 2020 Workshops: Glasgow, UK, August 23--28, 2020, Proceedings, Part II 16},
	pages={577--594},
	year={2020},
	organization={Springer}
}

@misc{driess2023palmeembodiedmultimodallanguage,
      title={PaLM-E: An Embodied Multimodal Language Model}, 
      author={Danny Driess and Fei Xia and Mehdi S. M. Sajjadi and Corey Lynch and Aakanksha Chowdhery and Brian Ichter and Ayzaan Wahid and Jonathan Tompson and Quan Vuong and Tianhe Yu and Wenlong Huang and Yevgen Chebotar and Pierre Sermanet and Daniel Duckworth and Sergey Levine and Vincent Vanhoucke and Karol Hausman and Marc Toussaint and Klaus Greff and Andy Zeng and Igor Mordatch and Pete Florence},
      year={2023},
      eprint={2303.03378},
      archivePrefix={arXiv},
      primaryClass={cs.LG},
      url={https://arxiv.org/abs/2303.03378}, 
}

@misc{liang2023codepolicieslanguagemodel,
      title={Code as Policies: Language Model Programs for Embodied Control}, 
      author={Jacky Liang and Wenlong Huang and Fei Xia and Peng Xu and Karol Hausman and Brian Ichter and Pete Florence and Andy Zeng},
      year={2023},
      eprint={2209.07753},
      archivePrefix={arXiv},
      primaryClass={cs.RO},
      url={https://arxiv.org/abs/2209.07753}, 
}

@misc{vemprala2023chatgptroboticsdesignprinciples,
      title={ChatGPT for Robotics: Design Principles and Model Abilities}, 
      author={Sai Vemprala and Rogerio Bonatti and Arthur Bucker and Ashish Kapoor},
      year={2023},
      eprint={2306.17582},
      archivePrefix={arXiv},
      primaryClass={cs.AI},
      url={https://arxiv.org/abs/2306.17582}, 
}

@misc{li2024visionlanguagefoundationmodelseffective,
      title={Vision-Language Foundation Models as Effective Robot Imitators}, 
      author={Xinghang Li and Minghuan Liu and Hanbo Zhang and Cunjun Yu and Jie Xu and Hongtao Wu and Chilam Cheang and Ya Jing and Weinan Zhang and Huaping Liu and Hang Li and Tao Kong},
      year={2024},
      eprint={2311.01378},
      archivePrefix={arXiv},
      primaryClass={cs.RO},
      url={https://arxiv.org/abs/2311.01378}, 
}

@misc{wang2025robobertendtoendmultimodalrobotic,
      title={RoboBERT: An End-to-end Multimodal Robotic Manipulation Model}, 
      author={Sicheng Wang and Sheng Liu and Weiheng Wang and Jianhua Shan and Bin Fang},
      year={2025},
      eprint={2502.07837},
      archivePrefix={arXiv},
      primaryClass={cs.RO},
      url={https://arxiv.org/abs/2502.07837}, 
}

@misc{kim2024openvlaopensourcevisionlanguageactionmodel,
      title={OpenVLA: An Open-Source Vision-Language-Action Model}, 
      author={Moo Jin Kim and Karl Pertsch and Siddharth Karamcheti and Ted Xiao and Ashwin Balakrishna and Suraj Nair and Rafael Rafailov and Ethan Foster and Grace Lam and Pannag Sanketi and Quan Vuong and Thomas Kollar and Benjamin Burchfiel and Russ Tedrake and Dorsa Sadigh and Sergey Levine and Percy Liang and Chelsea Finn},
      year={2024},
      eprint={2406.09246},
      archivePrefix={arXiv},
      primaryClass={cs.RO},
      url={https://arxiv.org/abs/2406.09246}, 
}

@misc{song2022denoisingdiffusionimplicitmodels,
      title={Denoising Diffusion Implicit Models}, 
      author={Jiaming Song and Chenlin Meng and Stefano Ermon},
      year={2022},
      eprint={2010.02502},
      archivePrefix={arXiv},
      primaryClass={cs.LG},
      url={https://arxiv.org/abs/2010.02502}, 
}

@INPROCEEDINGS{9578682,
  author={Wang, Gu and Manhardt, Fabian and Tombari, Federico and Ji, Xiangyang},
  booktitle={2021 IEEE/CVF Conference on Computer Vision and Pattern Recognition (CVPR)}, 
  title={GDR-Net: Geometry-Guided Direct Regression Network for Monocular 6D Object Pose Estimation}, 
  year={2021},
  volume={},
  number={},
  pages={16606-16616},
  doi={10.1109/CVPR46437.2021.01634}}

@ARTICLE{9722997,
  author={Chen, Long and Yang, Han and Wu, Chenrui and Wu, Shiqing},
  journal={IEEE Robotics and Automation Letters}, 
  title={MP6D: An RGB-D Dataset for Metal Parts’ 6D Pose Estimation}, 
  year={2022},
  volume={7},
  number={3},
  pages={5912-5919},
  doi={10.1109/LRA.2022.3154807}}

@misc{sun2025exploringposeguidedimitationlearning,
      title={Exploring Pose-Guided Imitation Learning for Robotic Precise Insertion}, 
      author={Han Sun and Yizhao Wang and Zhenning Zhou and Shuai Wang and Haibo Yang and Jingyuan Sun and Qixin Cao},
      year={2025},
      eprint={2505.09424},
      archivePrefix={arXiv},
      primaryClass={cs.RO},
      url={https://arxiv.org/abs/2505.09424}, 
}

@INPROCEEDINGS{9340842,
  author={Sock, Juil and Garcia-Hernando, Guillermo and Kim, Tae-Kyun},
  booktitle={2020 IEEE/RSJ International Conference on Intelligent Robots and Systems (IROS)}, 
  title={Active 6D Multi-Object Pose Estimation in Cluttered Scenarios with Deep Reinforcement Learning}, 
  year={2020},
  volume={},
  number={},
  pages={10564-10571},
  doi={10.1109/IROS45743.2020.9340842}}

@misc{yang2025active6dposeestimation,
      title={Active 6D Pose Estimation for Textureless Objects using Multi-View RGB Frames}, 
      author={Jun Yang and Wenjie Xue and Sahar Ghavidel and Steven L. Waslander},
      year={2025},
      eprint={2503.03726},
      archivePrefix={arXiv},
      primaryClass={cs.CV},
      url={https://arxiv.org/abs/2503.03726}, 
}

@misc{shao2025largevlmbasedvisionlanguageactionmodels,
      title={Large VLM-based Vision-Language-Action Models for Robotic Manipulation: A Survey}, 
      author={Rui Shao and Wei Li and Lingsen Zhang and Renshan Zhang and Zhiyang Liu and Ran Chen and Liqiang Nie},
      year={2025},
      eprint={2508.13073},
      archivePrefix={arXiv},
      primaryClass={cs.RO},
      url={https://arxiv.org/abs/2508.13073}, 
}

@misc{panerati2021learningflygym,
      title={Learning to Fly -- a Gym Environment with PyBullet Physics for Reinforcement Learning of Multi-agent Quadcopter Control}, 
      author={Jacopo Panerati and Hehui Zheng and SiQi Zhou and James Xu and Amanda Prorok and Angela P. Schoellig},
      year={2021},
      eprint={2103.02142},
      archivePrefix={arXiv},
      primaryClass={cs.RO},
      url={https://arxiv.org/abs/2103.02142}, 
}

@misc{he2020pvn3ddeeppointwise3d,
      title={PVN3D: A Deep Point-wise 3D Keypoints Voting Network for 6DoF Pose Estimation}, 
      author={Yisheng He and Wei Sun and Haibin Huang and Jianran Liu and Haoqiang Fan and Jian Sun},
      year={2020},
      eprint={1911.04231},
      archivePrefix={arXiv},
      primaryClass={cs.CV},
      url={https://arxiv.org/abs/1911.04231}, 
}

@misc{chi2024diffusionpolicyvisuomotorpolicy,
      title={Diffusion Policy: Visuomotor Policy Learning via Action Diffusion}, 
      author={Cheng Chi and Zhenjia Xu and Siyuan Feng and Eric Cousineau and Yilun Du and Benjamin Burchfiel and Russ Tedrake and Shuran Song},
      year={2024},
      eprint={2303.04137},
      archivePrefix={arXiv},
      primaryClass={cs.RO},
      url={https://arxiv.org/abs/2303.04137}, 
}

\end{document}